\documentclass{article}

\usepackage{microtype}
\usepackage{graphicx}
\usepackage{booktabs} 
\usepackage{amsfonts}
\usepackage{amsmath,amssymb}
\usepackage{amsthm}
\usepackage{footnote}
\usepackage{bbm}
\usepackage{subfig}
\usepackage{dblfloatfix}

\usepackage{hyperref}

\newcommand{\bilinear}[3]{\ensuremath{\vctr{z}_{\vertex{e}_{#1}}^T \mtrx{Z}_{\relation{r}_{#2}} \vctr{z}_{\vertex{e}_{#3}}}} 

\newcommand{\vctr}[1]{\ensuremath{\pmb{#1}}}
\newcommand{\mtrx}[1]{\ensuremath{\pmb{#1}}}
\newcommand{\graph}[1]{\ensuremath{\mathcal{#1}}}
\newcommand{\vertices}[1]{\ensuremath{\mathcal{#1}}}

\newcommand{\vertex}[1]{\ensuremath{\mathsf{#1}}}

\newcommand{\relations}[1]{\ensuremath{\mathcal{#1}}}
\newcommand{\model}[1]{\ensuremath{\mathcal{#1}}}

\newcommand{\relation}[1]{\ensuremath{\mathsf{#1}}}

\newcommand{\function}[1]{\ensuremath{\mathtt{#1}}}

\newtheorem{theorem}{Theorem}
\newtheorem{corollary}{Corollary}

\usepackage[accepted]{icml2020}

\icmltitlerunning{Knowledge Graph Embedding Without Negative Sampling}

\begin{document}

\twocolumn[
\icmltitle{Stay Positive: Knowledge Graph Embedding Without Negative Sampling}



\icmlsetsymbol{equal}{*}

\begin{icmlauthorlist}
\icmlauthor{Ainaz Hajimoradlou}{bor1}
\icmlauthor{Seyed Mehran Kazemi}{bor}
\end{icmlauthorlist}

\icmlaffiliation{bor}{BorealisAI, Montreal, Canada}
\icmlaffiliation{bor1}{BorealisAI, Vancouver, Canada}

\icmlcorrespondingauthor{Ainaz Hajimoradlou}{ainaz.hajimoradlou@borealisai.com}

\icmlkeywords{Machine Learning, ICML}

\vskip 0.3in
]



\printAffiliationsAndNotice{}  

\begin{abstract}
Knowledge graphs (KGs) are typically incomplete and we often wish to infer new facts given the existing ones. This can be thought of as a binary classification problem; we aim to predict if new facts are true or false. Unfortunately, we generally only have positive examples (the known facts) but we also need negative ones to train a classifier. To resolve this, it is usual to generate negative examples using a {\em negative sampling} strategy. However, this can produce false negatives which may reduce performance, is computationally expensive, and does not produce calibrated classification probabilities. In this paper, we propose a training procedure that obviates the need for negative sampling by adding a novel regularization term to the loss function. Our results for two relational embedding models (DistMult and SimplE) show the merit of our proposal both in terms of performance and speed.
\end{abstract}

\section{Introduction} \label{sec:intro}
Relational embedding models have proved effective for link prediction in knowledge graphs (KGs) \cite{nickel2016review,nguyen2017overview,wang2017knowledge,kazemi2020relational}. 
These models learn a mapping from each entity $\vertex{e}\in\vertices{E}$ and each relation $\relation{r}\in\relations{R}$ to a representation in a latent space such that a pre-defined score function $\phi$ operating in that latent space can make a prediction about whether a triple $(\vertex{h},\relation{r},\vertex{t})$ is true or not based on the representations for $\vertex{h}$, $\relation{r}$, and $\vertex{t}$. 

Ideally, if a KG contained a set of facts and a set of falsehoods (corresponding to positive and negative examples), one could formulate link prediction as a binary classification problem. However, a particular problem in training link prediction models is that KGs only contain facts (i.e., positive examples). To address this problem, link prediction models resort to negative sampling techniques to generate a set of triples that are expected to have a low chance of being true and use those triples as negative examples.

Current negative sampling strategies have been criticized for several reasons including generating large numbers of false-negatives, generating negative examples that are far from the model's decision boundary and produce near-zero loss, and making the models produce uncalibrated probabilities \cite{wang2018incorporating,shan2018confidence}. Moreover, as evidenced in Table~\ref{tab:timing} for two sample models (ComplEx \cite{trouillon2016complex} and SimplE \cite{kazemi2018simple}\footnote{The timings are for the FB15k dataset \cite{bollacker2008freebase} using the best hyperparameters reported for these models. For ComplEx, we used the implementation provided by the authors at \url{https://github.com/ttrouill/complex}. For SimplE, we used the fast version available at \url{https://github.com/baharefatemi/SimplE}.}), a large portion of training time is typically devoted to generating negative samples. Using negative examples during training either requires a substantial increase in the size of mini-batches and consequently in the required amount of GPU memory or requires increasing the number of mini-batches which slows down training. This can cause issues for real-world KGs that are orders of magnitude larger than the benchmark datasets \cite{balkir2019using}.  Nevertheless, negative sampling remains the mainstream approach primarily due to the lack of better alternatives.

\begin{table}[t]
    \caption{The percentage of time spent on negative sampling and on the rest of the operations for each training epoch of SimplE and ComplEx on Fb15k. The results are averaged over 100 epochs. Note that besides the negative sampling column, the other column is also affected by negative sampling.}
    \label{tab:timing}
    \begin{tabular}{lccc}
    \toprule
        Model & Library & Negative Sampling & Rest \\ \hline
        SimplE & PyTorch & 69.3\% & 30.7\% \\
        ComplEx & Theano & 62.3\% & 37.7\% \\
    \end{tabular}
\end{table}

In this paper, we obviate the need for negative sampling through proposing a one-class classification framework where we train models  using only the positive examples and a novel regularization loss. 
As an orthogonal contribution, we identify issues with WN18RR \cite{dettmers2018convolutional} and FB15k-237 \cite{toutanova2015observed}, two well-established benchmarks for link prediction in KGs, and create cleaned versions that resolve the identified issue. 

\section{Background and Notation} \label{sec:background}
For a set $\vertices{E}$ of entities and $\relations{R}$ of relations, let $\zeta_{\vertices{E}, \relations{R}}$ represent the set of all triples of the form $(\vertex{h}, \relation{r}, \vertex{t})$ such that $\vertex{h},\vertex{t}\in\vertices{E}$ and $\relation{r}\in\relations{R}$ where the triple is a fact. A \emph{knowledge graph (KG)} $\graph{G}_{\vertices{E}, \relations{R}}\subset\zeta_{\vertices{E}, \relations{R}}$ contains a subset of the facts in $\zeta_{\vertices{E}, \relations{R}}$. \emph{KG completion} is the problem of inferring $\zeta_{\vertices{E}, \relations{R}}$ from $\graph{G}_{\vertices{E}, \relations{R}}$. 
A relational embedding model $\model{M}(\theta)$ maps each entity and each relation to a hidden representation known as \emph{embedding} and defines a  function $\phi_{\model{M}(\theta)}:\vertices{E}\times\relations{R}\times\vertices{E} \rightarrow \mathbb{R}$ from the head and tail entities and the relation in a triple to a plausibility score. 
To convert the score into a probability, one can take the Sigmoid of the score.

\textbf{Bilinear Models:} are a successful class of embedding models. A bilinear model learns a vector embedding $\vctr{z}_\vertex{e}\in\mathbb{R}^d$ for each entity $\vertex{e}\in\vertices{E}$ and a matrix embedding $\mtrx{Z}_\vertex{r}\in\mathbb{R}^{d\times d}$ for each relation $\relation{r}\in\relations{R}$, where $d$ is the embedding dimension. The score function for a triple $(\vertex{h}, \relation{r}, \vertex{t})$ is then defined as $\phi_{\model{M}(\theta)}(\vertex{h}, \relation{r}, \vertex{t}) = \vctr{z}_\vertex{h}^T\mtrx{Z}_\relation{r}\vctr{z}_\vertex{t}$.
Bilinear models mainly differ in the restrictions they impose on the $\mtrx{Z}_\relation{r}$ matrices.

\textbf{Training:} Let $\graph{G}_{\vertices{E}, \relations{R}}$ be a KG and $\model{M}(\theta)$ be a model. The parameters $\theta$ are typically learned by minimizing a loss function similar to the following: 
\begin{multline}
 \label{eq:negsamp-loss}
\mathcal{L(\theta)} = \sum_{(\vertex{h},\relation{r},\vertex{t})\in \graph{G}_{\vertices{E}, \relations{R}}}\Big( \mathcal{L^{+}\big(\phi_{\model{M}(\theta)}(\vertex{h},\relation{r},\vertex{t})\big)}~+\\
\sum_{(\vertex{h'},\relation{r'},\vertex{t'})\in \function{Neg}(\vertex{h},\relation{r},\vertex{t};~n)} \mathcal{L^{-}}\big(\phi_{\model{M}(\theta)}(\vertex{h'},\relation{r'},\vertex{t'})\big)\Big)
\end{multline}
where $\function{Neg}(\vertex{h},\relation{r},\vertex{t}; n)$ is a negative sampling function that produces $n$ negative examples for the given positive example $(\vertex{h},\relation{r},\vertex{t})$ and $\mathcal{L}^+$ and $\mathcal{L}^-$ are the loss functions for positive and negative triples respectively. The hyperparameter $n$ is called the \emph{negative ratio}. A common choice for $\mathcal{L}^+$ and $\mathcal{L}^-$, also used in our experiments, is the negative log-likelihood: $\mathcal{L}^l\big(\phi_{\model{M}(\theta)}(\vertex{h},\relation{r},\vertex{t})\big)=\function{softplus}\big(-l*\phi_{\model{M}(\theta)}(\vertex{h},\relation{r},\vertex{t})\big)$.

\section{An Alternative to Negative Sampling} \label{sec:method}
We define our loss function as:
\begin{equation} \label{eq:loss-reg}
\mathcal{L(\theta)} = \sum_{(\vertex{h},\relation{r},\vertex{t})\in \graph{G}_{\vertices{E}, \relations{R}}} \mathcal{L}^{+}(\phi_{\model{M}(\theta)}(\vertex{h},\relation{r},\vertex{t})) + \lambda \mathcal{L}^{sp}(\phi_{\model{M}(\theta)})
\end{equation}
where the first term aims at penalizing low scores for the positive examples in $\graph{G}_{\vertices{E}, \relations{R}}$, $\mathcal{L}^{sp}(\phi_{\model{M}(\theta)})$ aims at preventing the model from generating high scores for all triples as well as giving the model a prior probability about the correctness of triples ($sp$ stands for \emph{stay positive}), and $\lambda$ is a hyperparameter.  
We define the regularization term as:
\begin{equation} \label{eq:reg}
\mathcal{L}^{sp}(\phi_{\model{M}(\theta)}) = \Big|\Big| \sum_{\vertex{h}\in \vertices{E}} \sum_{\relation{r}\in \relations{R}} \sum_{\vertex{t}\in\vertices{E}} \phi_{\model{M}(\theta)}(\vertex{h},\relation{r},\vertex{t}) - \psi |\vertices{E}|^2|\relations{R}| \Big|\Big|_p
\end{equation}
$\mathcal{L}^{sp}$ encourages the sum of the model scores for all triples to be close to $\psi |\vertices{E}|^2|\relations{R}|$ (or equivalently the average to be close to $\psi$)\footnote{In each batch, we compute the regularization term only for entities and relations in the batch; computing it for all entities and relations did not offer an improvement in our experiments.}. By encouraging the average score to be around $\psi$, for a randomly constructed triple $(\vertex{h}, \relation{r}, \vertex{t})$ we have $\mathbb{E}[\phi_{\model{M}(\theta)}(\vertex{h}, \relation{r}, \vertex{t})]\approx \psi$. Therefore, $\sigma(\psi)$ can be considered as the prior probability of the model about the correctness of triples (recall that $\phi$ provides a score and $\sigma(\phi)$ provides a probability). If besides the positive samples in the KG we also know (an approximation of) the ratio of positive to negative triples (corresponding to the prior probability of the correctness of triples), we can set $\psi$ accordingly. Otherwise, we consider $\psi$ as a hyperparameter.

In practice, we observed that optimizing the sum of all scores in Equation~\eqref{eq:reg} to be close to zero is easier than optimizing it to be close to a non-zero number. To take this observation into account, instead of optimizing for a model $\model{M}(\theta)$, we optimize another model $\model{M'}(\theta)$ where $\phi_{\model{M'}(\theta)}=\phi_{\model{M}(\theta)}+\psi$. Then, Equation~\eqref{eq:loss-reg} can be re-written for $\model{M'}(\theta)$ as
$
\mathcal{L(\theta)} = \sum_{(\vertex{h},\relation{r},\vertex{t})\in \graph{G}_{\vertices{E}, \relations{R}}} \mathcal{L}^{+}(\phi_{\model{M}(\theta)}(\vertex{h},\relation{r},\vertex{t}) + \psi)
+ \lambda \mathcal{L}^{sp}(\phi_{\model{M'}(\theta)})$
and the regularization term can be re-written as
$\mathcal{L}^{sp}(\phi_{\model{M}(\theta)}) = \Big|\Big| \sum_{\vertex{h}\in \vertices{E}} \sum_{\relation{r}\in \relations{R}} \sum_{\vertex{t}\in\vertices{E}} \phi_{\model{M}(\theta)}(\vertex{h},\relation{r},\vertex{t}) \Big|\Big|_p
$.

\textbf{Regularization vs closed world assumption (CWA):} CWA (corresponding to assuming any triple that is not in the KG is false) is a specific case of our model with $\psi$ set to $\frac{|\graph{G}_{\vertices{E},\relation{R}}|}{|\vertices{E}|^2|\relations{R}|}$, where $|\graph{G}_{\vertices{E},\relation{R}}|$ corresponds to the number of facts in the KG. But our regularization is more general in that as $\psi$ becomes larger, the CWA becomes more relaxed. Unlike negative sampling, training with our regularization term never asks the model to provide a negative score for a specific triple. Instead, it provides the model with the flexibility of assigning any score to any triple as long as the mean of the scores matches with $\psi$. 

\textbf{Time Complexity:} While a naive calculation of Equation~\eqref{eq:reg} requires $O(|\vertices{E}|^2|\relations{R}|)$ score computations we prove that for bilinear models it can be computed efficiently.

\begin{theorem}
Let $\model{M}$ be a bilinear model with embedding dimension $d$ where for each relation, only $d \leq \alpha \leq d^2$ of the elements in the embedding matrix can be non-zero. For a knowledge graph $\graph{G}_{\vertices{E}, \relations{R}}$, the sum of  scores of $\model{M}$ for all possible triples can be computed in $O\big(|\mathcal{E}|d+|\mathcal{R}|\alpha\big)$ as $(\sum_{\vertex{e}\in\vertices{E}}  \vctr{z}_\vertex{e}^T) (\sum_{\relation{r}\in\relations{R}}  \mtrx{Z}_\relation{r}) (\sum_{\vertex{e}\in\vertices{E}}  \vctr{z}_\vertex{e})$.
\end{theorem}

\begin{corollary} \label{cor-rescal}
Since $\alpha\in O(d^2)$ for RESCAL, for a knowledge graph $\graph{G}_{\vertices{E}, \relations{R}}$, the sum of the RESCAL scores for all possible triples can be computed in $O\big(|\mathcal{E}|d+|\mathcal{R}|d^2\big)$.
\end{corollary}

Corollary~\ref{cor-rescal} can be extended to the models that use a combination of bilinear score functions such as neural tensor networks \cite{socher2013reasoning}.

\begin{corollary}
Since $\alpha\in O(d)$ in DistMult, ANALOGY, SimplE, ComplEx, HolE (and several other bilinear models), for a knowledge graph $\graph{G}_{\vertices{E}, \relations{R}}$, the sum of the model scores for all possible triples for any of the above models can be computed in $O\big((|\mathcal{E}|+|\mathcal{R}|)d\big)$.
\end{corollary}

For a KG $\graph{G}_{\vertices{E}, \relations{R}}$, let $\beta$ represent the average number of triples each entity in $\vertices{E}$ appears in. Let $n$ represent the negative ratio. Let $\model{M}$ be a bilinear model with embedding dimension $d$ where for each relation, only $\alpha$ of the elements in the embedding matrix can be non-zero. Assume $\alpha \geq d$ (this is true for all models mentioned so far) and $|\vertices{E}| > |\relations{R}|$ (in general and in each batch).
The following theorems establish the time complexity of a single training epoch using Equation~\ref{eq:negsamp-loss} and Equation~\ref{eq:loss-reg}.

\begin{theorem} \label{theorem-epoch-neg}
The time complexity of a single training epoch using Equation~\ref{eq:negsamp-loss} is $O(\beta (1+n)|\vertices{E}|\alpha)$.
\end{theorem}

\begin{theorem} \label{theorem-epoch-sp}
The time complexity of a single training epoch using Equation~\ref{eq:loss-reg} is $O(\beta |\vertices{E}|\alpha)$.
\end{theorem}

Theorems~\ref{theorem-epoch-neg}~and~\ref{theorem-epoch-sp} indicate that the time complexity of a single epoch when training with our regularization term is $(1+n)$ times less than training with negative sampling. 

\textbf{A practical consideration:} In Equation~\eqref{eq:reg}, a few scores may become extremely large and affect the total sum in an undesired way. For this reason, in our experiments, for the models trained with the regularization in Equation~\eqref{eq:reg}, we restrict the elements of the embedding vectors to be between $-1$ and $1$ (by applying an element-wise \function{Tanh} function) resulting in $\phi_\model{M}\in(-d, d)$ where $d$ is the embedding dimension. Then we multiply the model score by $\frac{I}{d}$, where $I$ is a hyperparameter, leading to $\phi_\model{M}\in(-I, I)$. 

\section{Experiments and Results}
We experiment with DistMult and SimplE as two representative bilinear models and provide head-to-head comparisons between their original training and the proposed training in this paper. We use \emph{spDistMult} and \emph{spSimplE} to refer to DistMult and SimplE trained using our regularization term, where \emph{sp} stands for \emph{Stay Positive}.

\textbf{Datasets:} Two established benchmarks for link prediction in KGs are \emph{WN18RR} \cite{dettmers2018convolutional} and \emph{FB15k-237} \cite{toutanova2015observed}. 
We noticed that a large number of entities that appear in validation and test sets of these two datasets do not appear in any training triples. For example, WN18RR contains 209 unseen entities in the test set. This is problematic because the predictions for these entities is mainly a function of initialization. As an independent contribution, we clean these two datasets further by removing the triples from validation and test sets involving unseen entities. We call the new datasets \emph{WN18AM} and \emph{FB15k-237AM}. WN18AM contains 40,559 entities, 11 relations, 86,835 train triples, 2,824 validation triples and 2,924 test triples. FB15k-237AM contains 14,505 entities, 237 relations, 272,115 train triples, 17,526 validation triples and 20,438 test triples.
We use the two new datasets as well as two other standard benchmarks that contain negative samples in their test sets so we can compare model probability calibrations. The two datasets are WN11 and FB13 both introduced in \cite{socher2013reasoning}. 

\begin{table*}[t]
\scriptsize
\caption{Results on four datasets. The columns under ``Original'' demonstrate the results of the model without any post-processing and the columns under ``Post-Calibrated'' demonstrate the results after post-calibrating the models using Platt scaling.}
\label{tab:nll_wn11_fb13}
\begin{center}
\resizebox{\textwidth}{!}{%
\begin{tabular}{ccccccccccccccc}
\toprule
& \multicolumn{5}{c}{WN11} & \multicolumn{5}{c}{FB13} & \multicolumn{2}{c}{WN18AM} & \multicolumn{2}{c}{FB15k-237AM}\\
\cmidrule(lr){2-6} \cmidrule(lr){7-11}
\cmidrule(lr){12-13}
\cmidrule(lr){14-15}
& \multicolumn{3}{c}{Original} & \multicolumn{2}{c}{Post-Calibrated} & \multicolumn{3}{c}{Original} & \multicolumn{2}{c}{Post-Calibrated} & & &  \\
\cmidrule(lr){2-4} \cmidrule(lr){5-6} \cmidrule(lr){7-9} \cmidrule(lr){10-11}
Model & NLL & Brier & AUC & NLL & Brier & NLL & Brier & AUC & NLL & Brier & MRR & Hit@1 & MRR & Hit@1 \\ \hline
DistMult & 2.670 & 0.430 & 0.668 & 0.675 & 0.241 & 1.572 & 0.393 & 0.695 & 0.630 & 0.219 & 0.440 & 0.410 & \textbf{0.256} & 0.160\\
SimplE & 2.385 & 0.418 & 0.692 & 0.676 & 0.242 & 1.504 & 0.388 & 0.661 & 0.619 & 0.214 & 0.461 & 0.256 & \textbf{0.256} & 0.163 \\
spDistMult & \textbf{0.620} & \textbf{0.218} & 0.701 & \textbf{0.582} & \textbf{0.202} & 0.679 & 0.243 & 0.759 & 0.568 & 0.189 & 0.448 & 0.423 & 0.249 & 0.181 \\
spSimplE & \textbf{0.620} & \textbf{0.218} & \textbf{0.710} & 0.586 & 0.205 & \textbf{0.589} & \textbf{0.201} & \textbf{0.793} & \textbf{0.556} & \textbf{0.184} & 
\textbf{0.470} & \textbf{0.445} & 0.250 & \textbf{0.184} \\
\end{tabular}
}
\end{center}
\end{table*}

\begin{figure*}[b]
\captionsetup[subfigure]{labelformat=empty}
   \centering
\subfloat[]{%
   \includegraphics[width=0.18\textwidth]{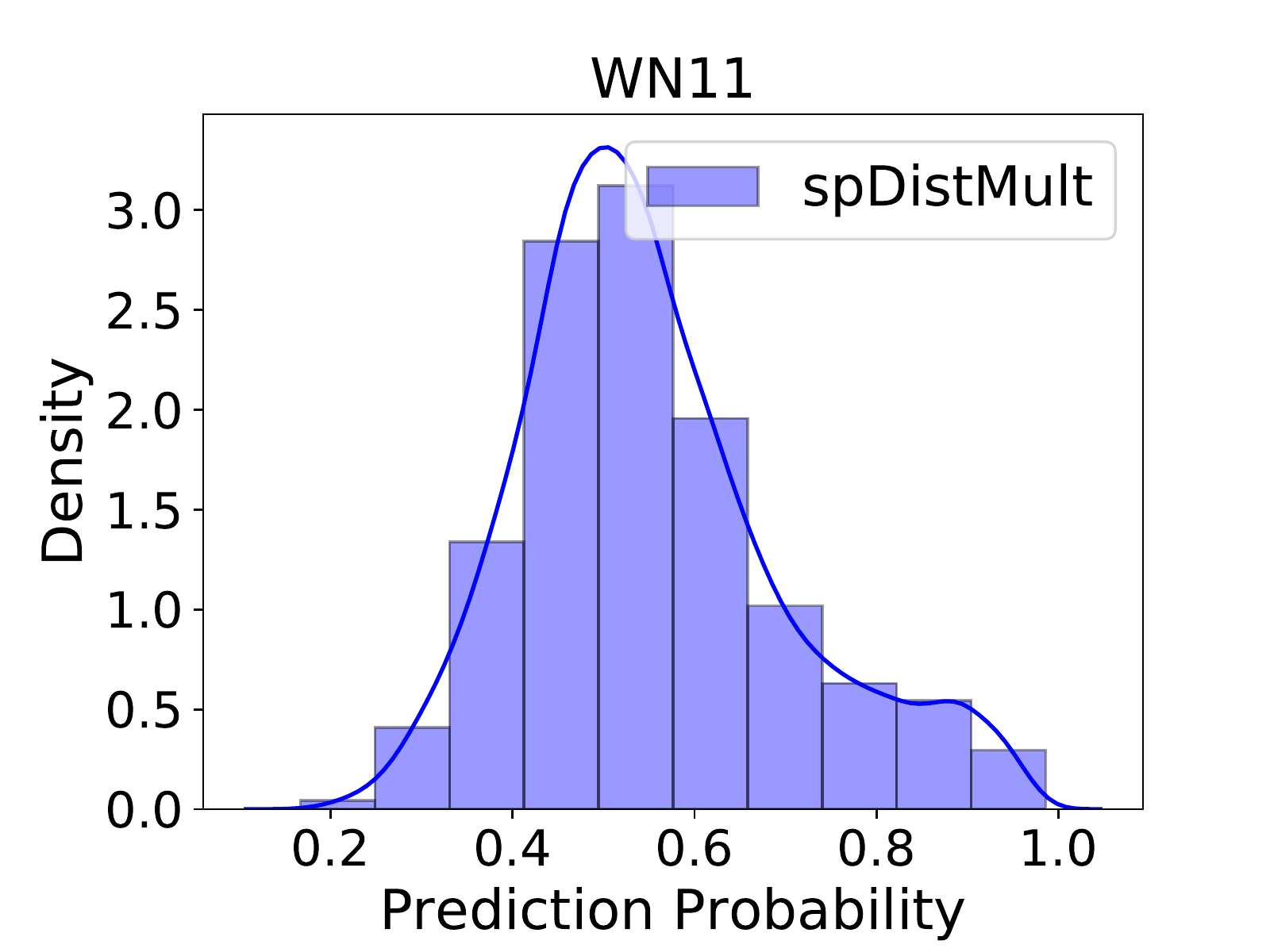}}
~~~~\hspace*{0cm}
   \subfloat[]{%
   \includegraphics[width=0.18\textwidth]{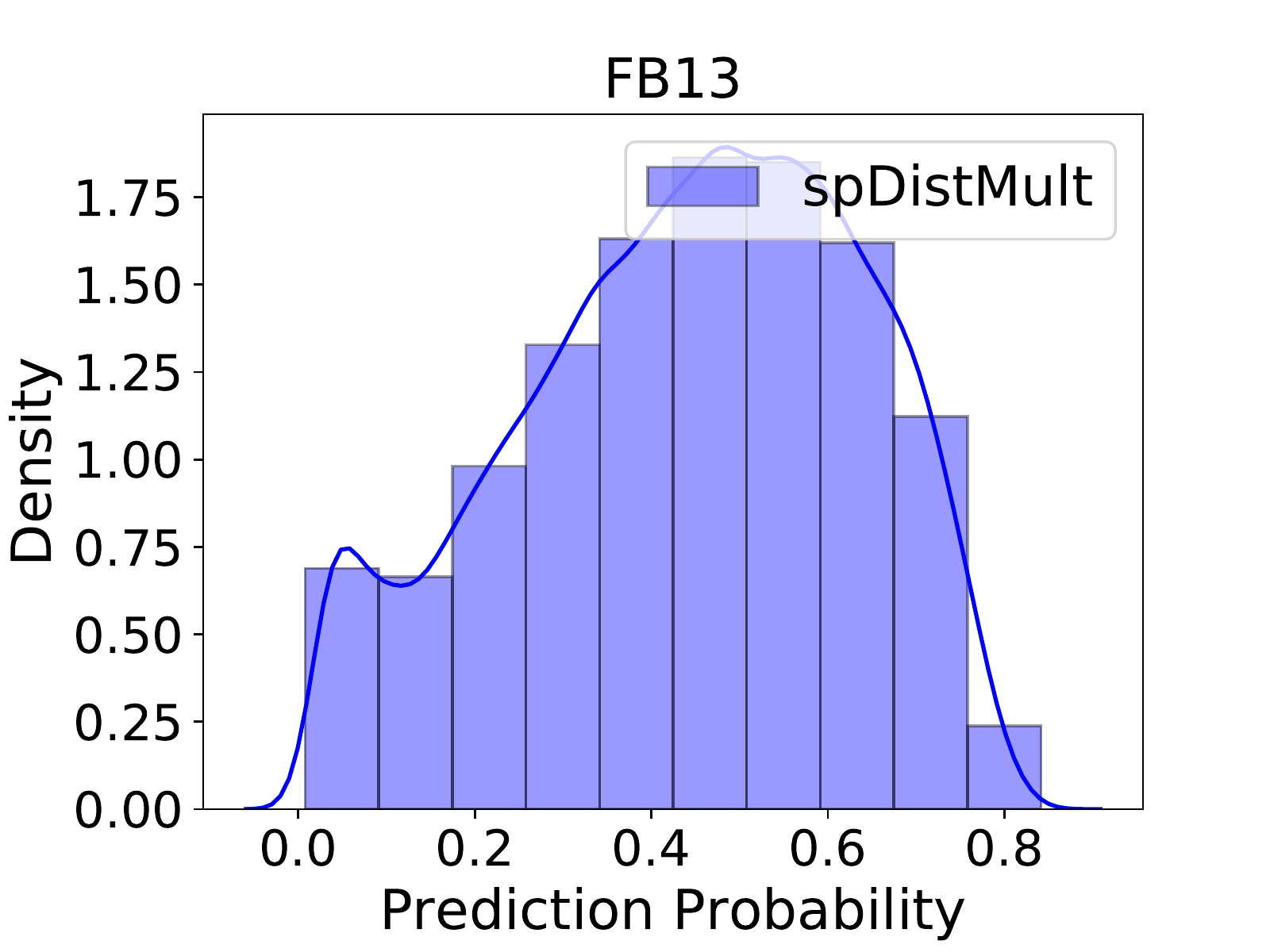}}
~~~~\hspace*{0cm}
   \subfloat[]{%
   \includegraphics[width=0.18\textwidth]{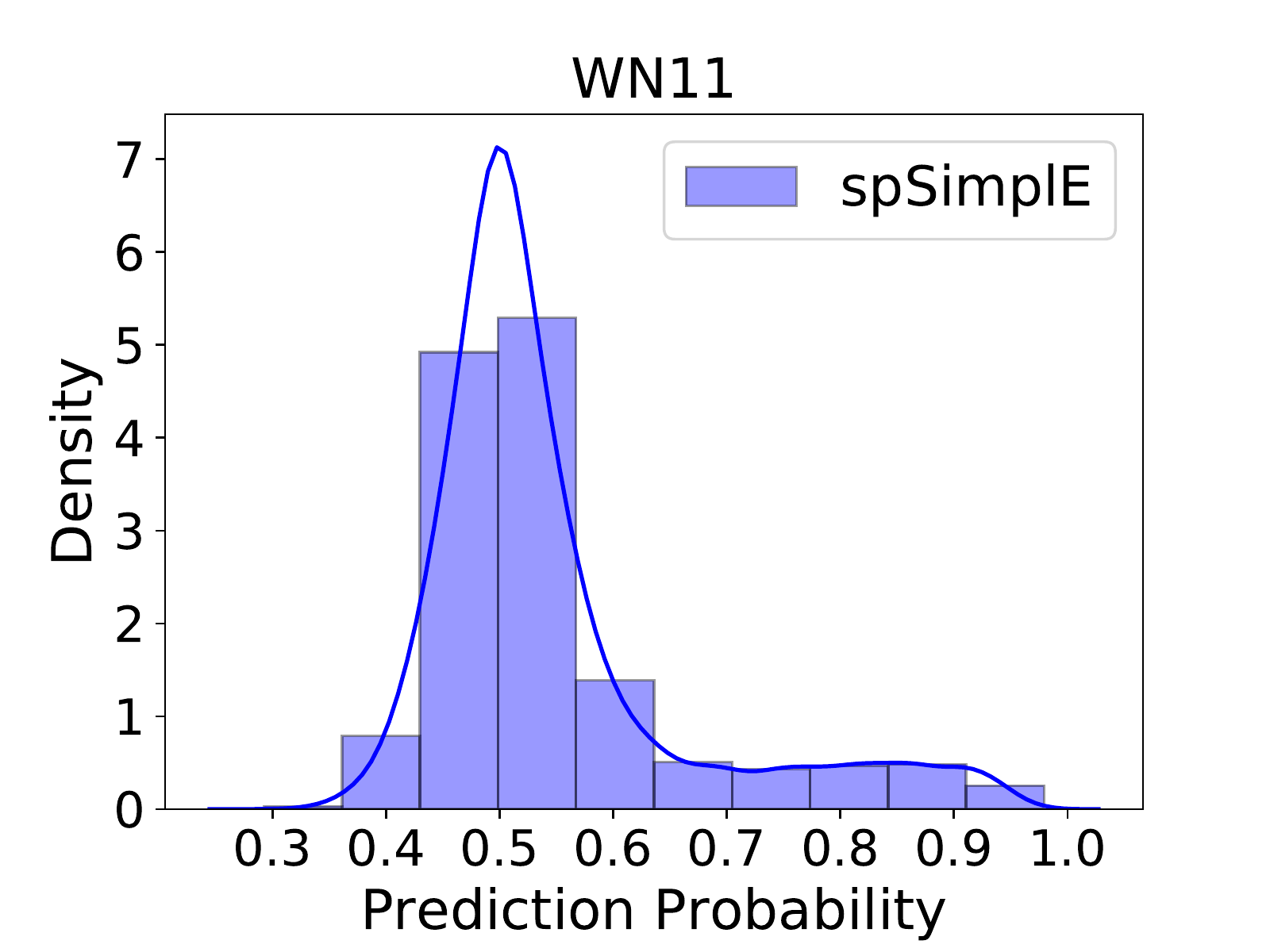}}
~~~~\hspace*{0cm}
   \subfloat[]{%
   \includegraphics[width=0.18\textwidth]{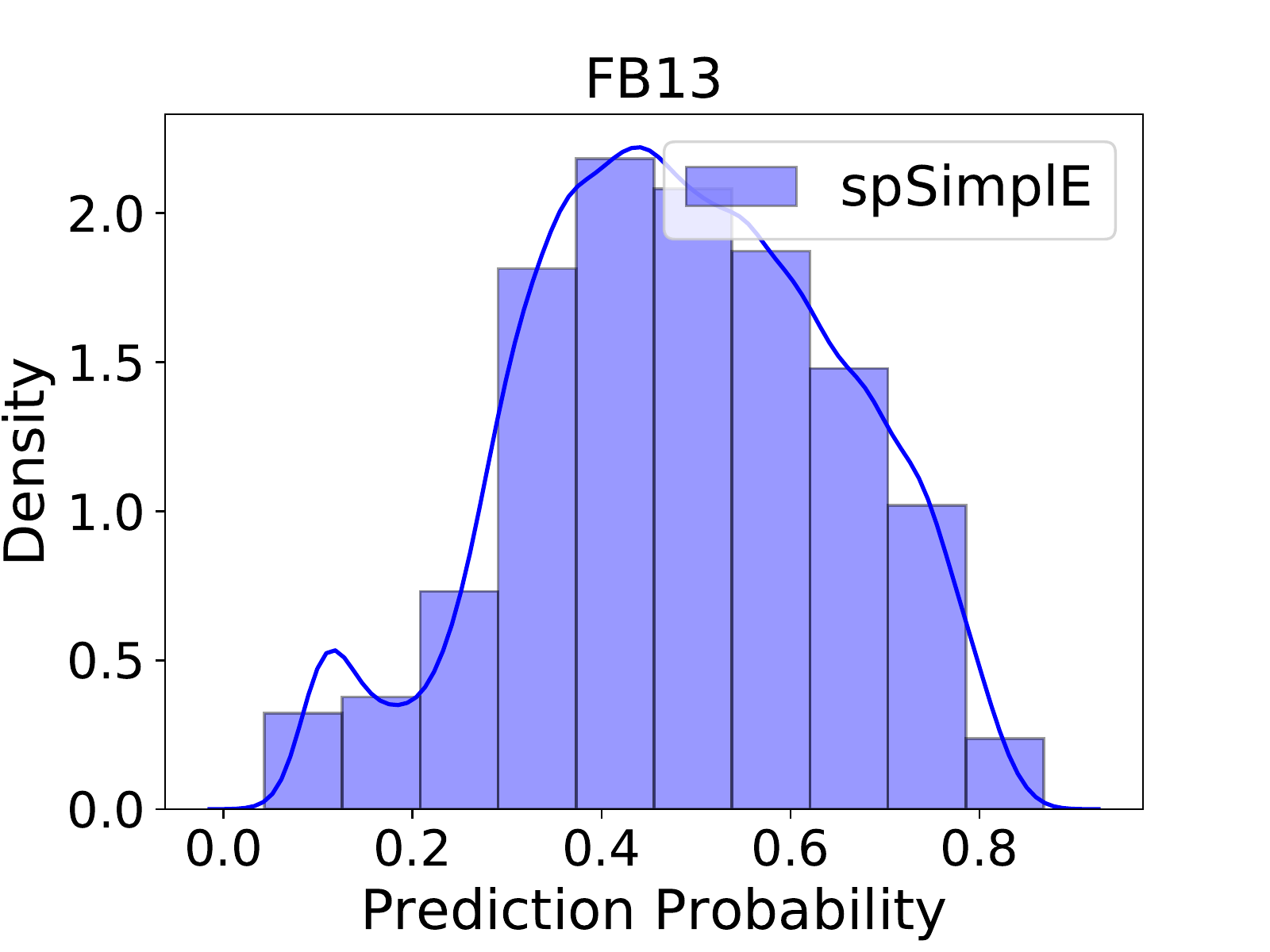}}
   \vspace{-0.85cm}
   \subfloat[]{%
   \includegraphics[width=0.18\textwidth]{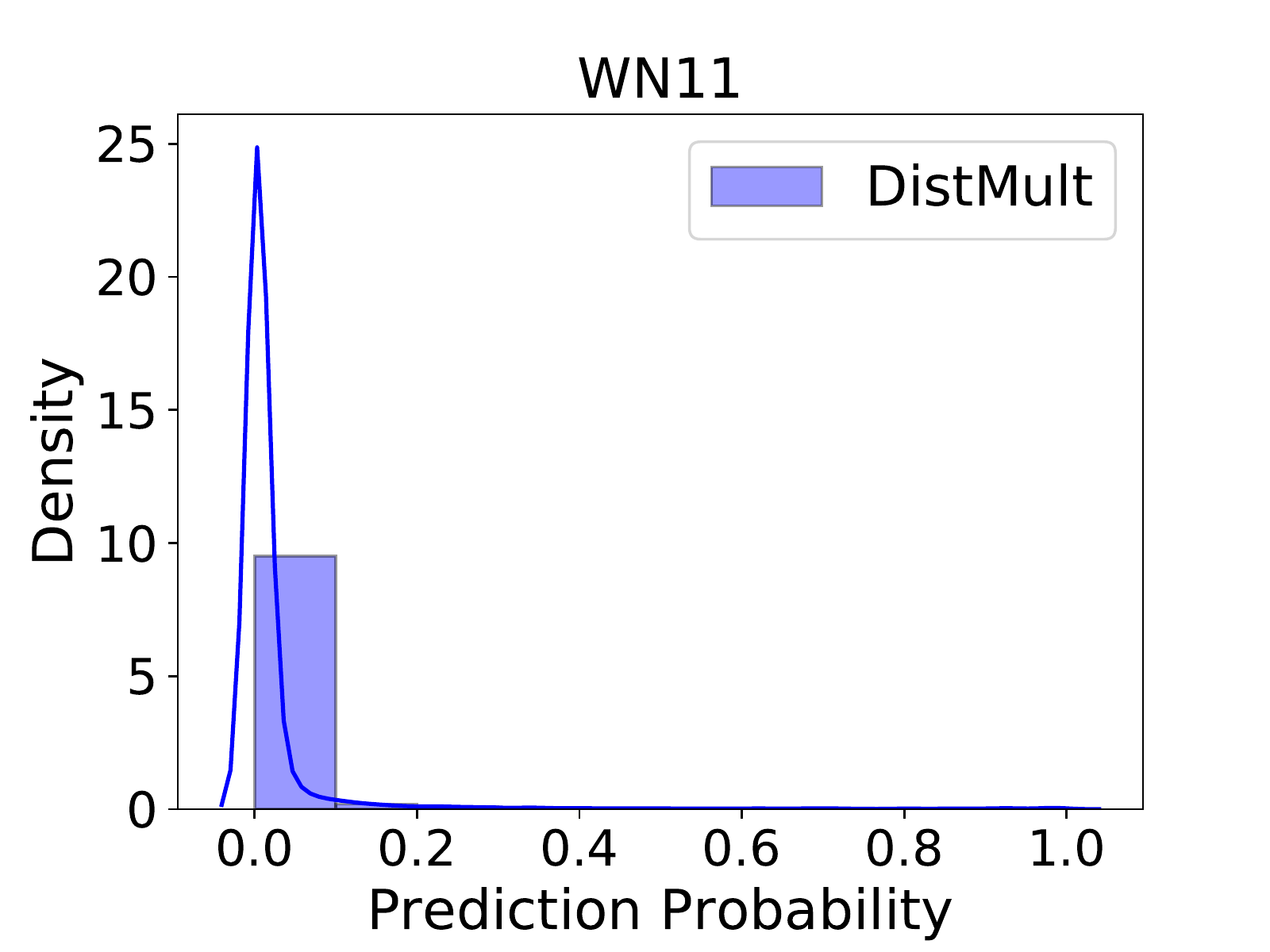}}
~~~~\hspace*{0cm}
   \subfloat[]{%
   \includegraphics[width=0.18\textwidth]{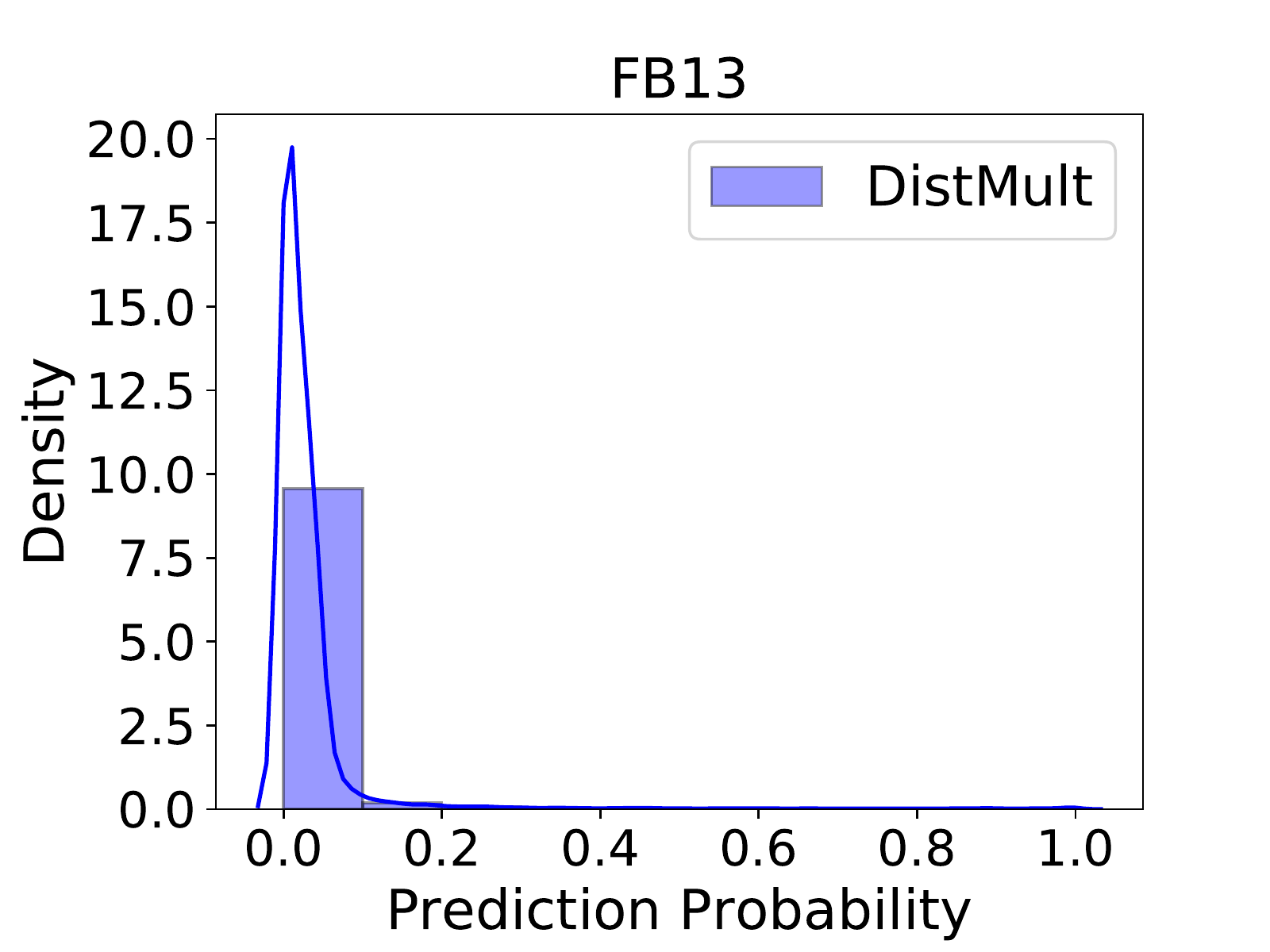}}
~~~~\hspace*{0cm}
   \subfloat[]{%
   \includegraphics[width=0.18\textwidth]{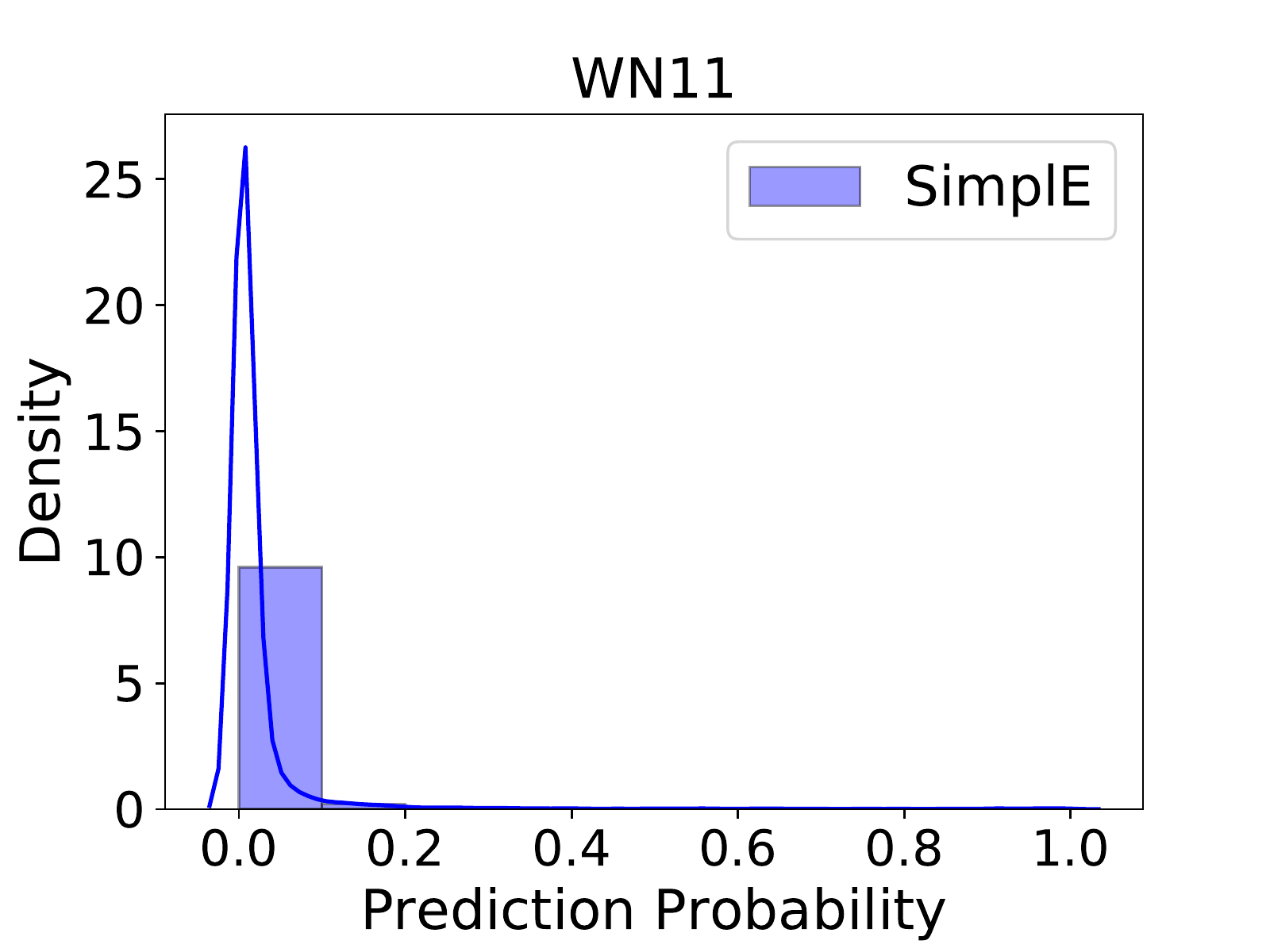}}
~~~~\hspace*{-0.1cm}
   \subfloat[]{%
   \includegraphics[width=0.18\textwidth]{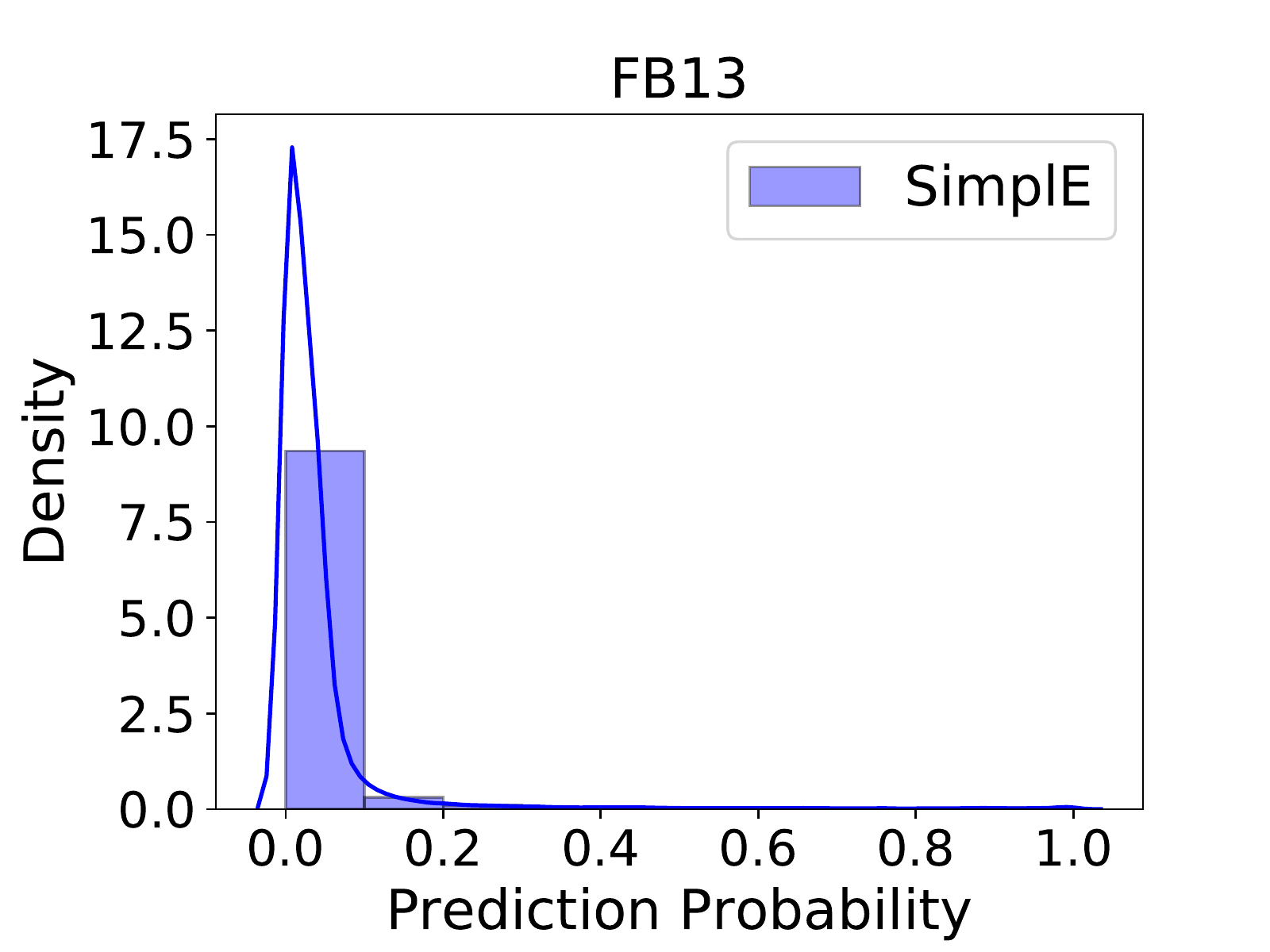}}

   \caption{%
   \label{fig:calib-diagrams} %
   The probability outputs of different models on WN11 and FB13 grouped into bins as well as the kernel density estimates.
   }
\end{figure*}

\textbf{Classification and calibration:} From the obtained results on WN11 and FB13 reported in Table~\ref{tab:nll_wn11_fb13} (the columns under ``Original''), one can see clear boosts for spDistMult and spSimplE in terms of NLL (up to $77\%$ improvement), Brier score (up to $49\%$), and also AUC (up to $20\%$).
In Figure~\ref{fig:calib-diagrams}, we bin the output probabilities of different models on the test sets of WN11 and FB13 and estimate a probability density function using kernel density estimation \cite{terrell1992variable}. It can be viewed that the output probabilities of DistMult and SimplE are mostly close to zero and squeezed in one bin; this explains their poor performance in terms of NLL and Brier score. These plots also match the plots of \citet{tabacof2020probability} for ComplEx. For spDistMult and spSimplE, however, it can be viewed that the output probabilities are more spread across bins and centered around 0.5. This explains (in part) the superior performance of spDistMult and spSimplE in terms of NLL and Brier score compared to DistMult and SimplE.

\textbf{Post-calibration:} \citet{tabacof2020probability} show empirically that post-calibration can substantially improve the output probabilities of the relational embedding models. Here, we examine how post-calibration affects our models and baselines. We use Platt scaling \cite{platt1999probabilistic} for post-calibration.
In Table~\ref{tab:nll_wn11_fb13} (``Post-Calibrated'' columns), we observe that post-calibration improves both the NLL and Brier score of the four models. Since DistMult and SimplE are originally highly uncalibrated, they are substantially impacted by post-calibration. For spDistMult and spSimplE, however, the impact is much smaller because they already produce better-calibrated outputs. Moreover, it can be viewed that even after post-calibration, in most cases DistMult and SimplE perform inferior compared to spDistMult and spSimplE \emph{without} post-calibration. Note that such post-calibrations are only possible if one has access to ground truth positive and negative examples in a validation set.

\begin{figure}[t]
    \centering
    \includegraphics[width=0.55\columnwidth]{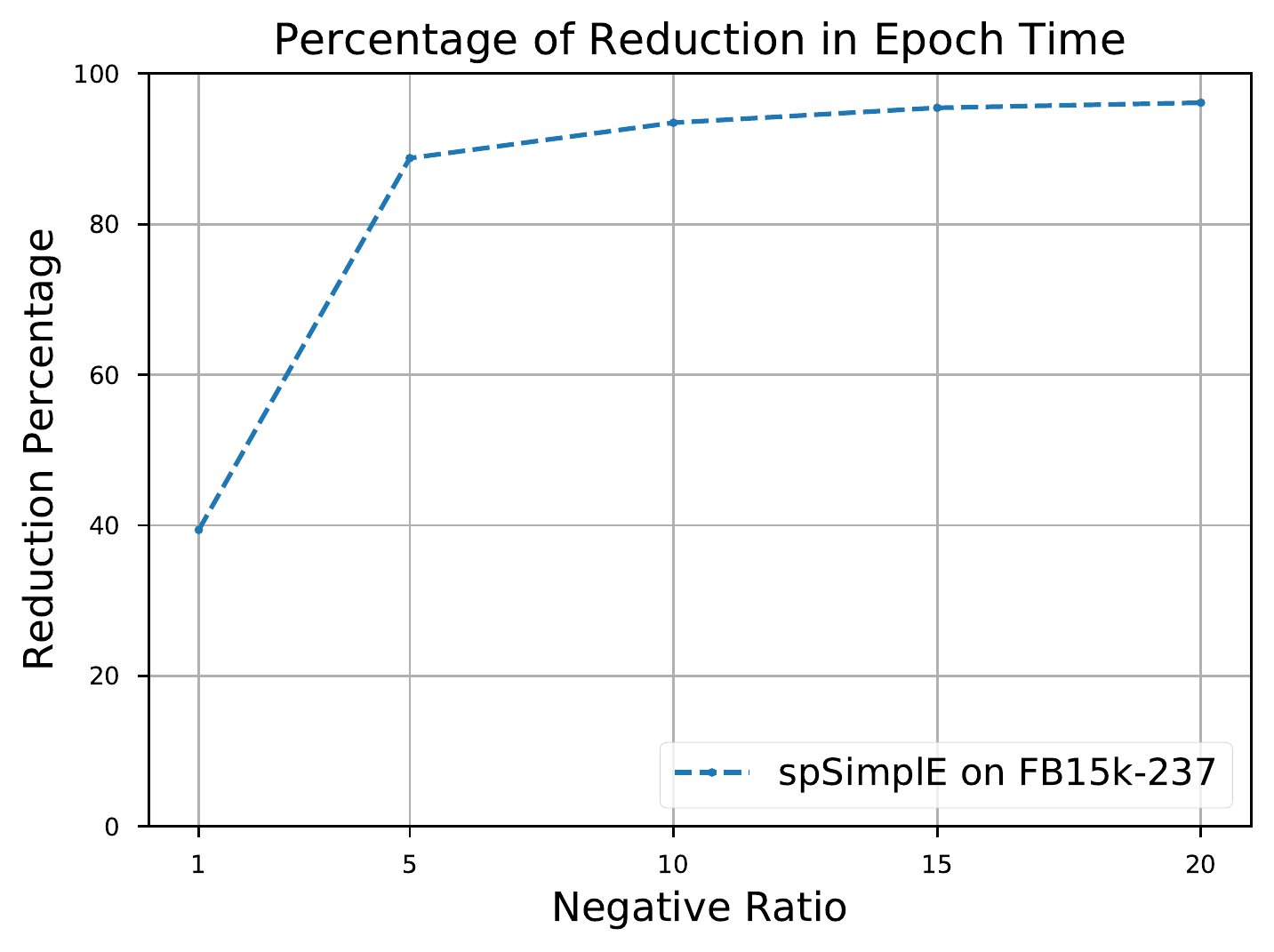}
    \caption{The percentage of reduction in epoch time achieved by spSimplE compared to SimplE for different negative ratios. The percentage is computed as $\frac{t_1-t_2}{t_1}$ where $t_1$ is the time for SimplE and $t_2$ is the time for spSimplE.}
    \label{fig:time-improvement}
\end{figure}

\textbf{Link prediction:} Table~\ref{tab:nll_wn11_fb13} represents the results obtained on WN18AM and FB15k-237AM. As can be viewed, without using any negative examples, spDistMult and spSimplE achieve comparable performance compared to DistMult and SimplE, showing that the regularization term can act as a replacement for negative samples. For further analysis, we measured the performance of the models and baselines when applying no dropout. We observed that the filtered MRR of DistMult reduced by $2.7\%$ and $18.4\%$ on WN18AM and FB15k-237AM respectively whereas this reduction for spDistMult is only $0.4\%$ and $0.8\%$. We got similar results for SimplE and spSimplE. These results show that the regularization term not only can act as a replacement for negative examples but also can have a regularization effect similar to that of dropout. Finally, we tested spDistMult and spSimplE when computing Equation~\ref{eq:reg} for all entities and relations as opposed to only entities and relations in the batch. We observed only minor changes to filtered MRR (less than $1\%$) showing that computing the regularization term only for entities and relations in the batch suffices.

\textbf{Running times:} As exemplified in Table~\ref{tab:timing}, negative sampling usually consumes a large portion of the time for each epoch. Here, we compare the running time of SimplE and spSimplE for each epoch on the FB15k-237AM dataset. For a fair comparison, for spSimplE we use a batch size $(n+1)$ times larger than that of SimplE as a rough estimate for memory usage (cf. Theorems~\ref{theorem-epoch-neg} and \ref{theorem-epoch-sp}). Figure~\ref{fig:time-improvement} shows the improvement achieved by spSimplE compared to SimplE on FB15k-237AM in terms of epoch time. The improvement is measured as the percentage of reduction in epoch time. It can be observed that spSimplE drastically reduces the epoch time especially for larger values of the negative ratio. 

\section{Discussion \& Conclusion} \label{sec:discussion}
In this paper, we proposed a novel regularization technique that obviates the need for negative examples during training and we showed its notable merits on several tasks and datasets. 
Future work can extend our regularization technique to other classes of models proposed in the literature, provide a deeper theoretical study of the proposed regularization and develop new regularization techniques, and extend the regularization to domains beyond relational embedding models where negative sampling is employed.

\bibliography{MyBib}
\bibliographystyle{icml2020}

\section*{Appendix}
\textbf{Proof of theorems:}
\setcounter{theorem}{0}
\begin{theorem}
Let $\model{M}$ be a bilinear model with embedding dimension $d$ where for each relation, only $d \leq \alpha \leq d^2$ of the elements in the embedding matrix can be non-zero. For a knowledge graph $\graph{G}_{\vertices{E}, \relations{R}}$, the sum of  scores of $\model{M}$ for all possible triples can be computed in $O\big(|\mathcal{E}|d+|\mathcal{R}|\alpha\big)$ as $(\sum_{\vertex{e}\in\vertices{E}}  \vctr{z}_\vertex{e}^T) (\sum_{\relation{r}\in\relations{R}}  \mtrx{Z}_\relation{r}) (\sum_{\vertex{e}\in\vertices{E}}  \vctr{z}_\vertex{e})$.
\end{theorem}

\begin{proof}
The sum of the scores of $\model{M}$ for all possible triples can be written as:
\begin{align*}
&~\bilinear{1}{1}{1} + \dots + \bilinear{1}{1}{|\vertices{E}|} + \\ &~\bilinear{1}{2}{1} + \dots + \bilinear{1}{2}{|\vertices{E}|} + \\
&~\dots + \\
&~ \bilinear{1}{|\relations{R}|}{1} + \dots + \bilinear{1}{|\relations{R}|}{|\vertices{E}|} + \\
&~ \dots + \\
&~\bilinear{|\vertices{E}|}{1}{1} + \dots + \bilinear{|\vertices{E}|}{1}{|\vertices{E}|} + \\ &~\bilinear{|\vertices{E}|}{2}{1} + \dots + \bilinear{|\vertices{E}|}{2}{|\vertices{E}|} + \\
&~\dots + \\
&~ \bilinear{|\vertices{E}|}{|\relations{R}|}{1} + \dots + \bilinear{|\vertices{E}|}{|\relations{R}|}{|\vertices{E}|}
\end{align*}
Since matrix product is (left) distributive with respect to matrix addition, for each row of the above expression, we can factor the first two elements of the bilinear score and re-write the sum as:
\begin{align*}
&~\vctr{z}_{\vertex{e}_{1}}^T \mtrx{Z}_{\relation{r}_1} (\sum_{\vertex{e} \in \vertices{E}} \vctr{z}_{\vertex{e}}) + \vctr{z}_{\vertex{e}_{1}}^T \mtrx{Z}_{\relation{r}_2} (\sum_{\vertex{e} \in \vertices{E}} \vctr{z}_{\vertex{e}}) +\dots +
\vctr{z}_{\vertex{e}_{1}}^T \mtrx{Z}_{\relation{r}_{|\relations{R}|}} (\sum_{\vertex{e} \in \vertices{E}} \vctr{z}_{\vertex{e}}) + \\
&~ \dots + \\
&~\vctr{z}_{\vertex{e}_{|\vertices{E}|}}^T \mtrx{Z}_{\relation{r}_1} (\sum_{\vertex{e} \in \vertices{E}} \vctr{z}_{\vertex{e}}) + \vctr{z}_{\vertex{e}_{|\vertices{E}|}}^T \mtrx{Z}_{\relation{r}_2} (\sum_{\vertex{e} \in \vertices{E}} \vctr{z}_{\vertex{e}}) +\dots +
\vctr{z}_{\vertex{e}_{|\vertices{E}|}}^T \mtrx{Z}_{\relation{r}_{|\relations{R}|}} (\sum_{\vertex{e} \in \vertices{E}} \vctr{z}_{\vertex{e}})
\end{align*}
Using the (right) distributivity of matrix product with respect to addition, for each row we can factor out the first vector of each element and re-write the above expression as:
\begin{align*}
(\sum_{\vertex{e} \in \vertices{E}} \vctr{z}^T_{\vertex{e}}) \Big( \mtrx{Z}_{\relation{r}_1} (\sum_{\vertex{e} \in \vertices{E}} \vctr{z}_{\vertex{e}}) + \dots + \mtrx{Z}_{\relation{r}_{|\relations{R}|}} (\sum_{\vertex{e} \in \vertices{E}} \vctr{z}_{\vertex{e}})\Big)
\end{align*}
Using the (right) distributivity of matrix product one more time for the expression inside the right parenthesis, we can re-write the above expression as:
\begin{align*}
    (\sum_{\vertex{e}\in\vertices{E}}  \vctr{z}_\vertex{e}^T) (\sum_{\relation{r}\in\relations{R}}  \mtrx{Z}_\relation{r}) (\sum_{\vertex{e}\in\vertices{E}}  \vctr{z}_\vertex{e})
\end{align*}
Computing the expression above requires computing two sums one over all entity vectors and one over all relation matrices. Summing over all entity vectors has a time complexity of $O(|\vertices{E}|d)$ and summing over all relation matrices has a time complexity of $O(|\relations{R}|\alpha)$ where $\alpha$ is the number of elements in relation matrices that are allowed to be non-zero. Therefore, the overall time complexity is $O\big(|\mathcal{E}|d+|\mathcal{R}|\alpha\big)$.
\end{proof}

\begin{theorem} \label{theorem-epoch-neg}
The time complexity of a single training epoch using Equation 1 in the main text is $O(\beta (1+n)|\vertices{E}|\alpha)$.
\end{theorem}

\begin{proof}
Assuming each entity appears on average in $\beta$ triples, a single epoch requires computing model scores for $O(\beta|\vertices{E}|)$ positive triples. With a negative ratio of $n$, a single epoch requires computing model scores for $O(\beta n|\vertices{E}|)$ triples. Therefore, overall a single epoch requires $O(\beta (1+n)|\vertices{E}|)$ score computations. Since the time complexity of computing a single model score is $O(\alpha)$, the overall time complexity becomes $O(\beta (1+n)|\vertices{E}|\alpha)$.
\end{proof}

\begin{theorem} \label{theorem-epoch-sp}
The time complexity of a single training epoch using Equation 2 in the main text is $O(\beta |\vertices{E}|\alpha)$.
\end{theorem}
\begin{proof}
To prove this theorem, we first need to look at each mini-batch. For a mini-batch $B$, let $\vertices{E}_B$ and $\relations{R}_B$ represent the entities and relations in the batch and $\beta_B$ represent the average number of triples each entity in $\vertices{E}_B$ appears in. With a similar reasoning as in the proof of Theorem~\ref{theorem-epoch-neg}, the time complexity for computing the score for the positive triples in this batch is $O(\beta_B |\vertices{E}_B|\alpha)$. The time complexity for computing the regularization term is $O(|\vertices{E}_B|d + |\vertices{R}_B|\alpha)$. Note that $\beta_B \geq 1$ because every entity in $|\vertices{E}_B|$ appears in at least one triple. We also assumed $|\vertices{E}_B| > |\relations{R}_B|$ and $\alpha \geq d$. Therefore, $|\vertices{E}_B|d < \beta_B |\vertices{E}_B|\alpha$ and $|\vertices{R}_B|\alpha < \beta_B |\vertices{E}_B|\alpha$ and so the time complexity for computing the regularization term gets subsumed into the time complexity of computing the model scores and we can ignore it. By summing the time complexities for computing the model scores for all batches, we obtain the overall time complexity of $O(\beta |\vertices{E}|\alpha)$.
\end{proof}

\textbf{Evaluation Metrics:} 
To measure the performance on WN11 and FB13, following \cite{tabacof2020probability}, we report negative log-likelihood (NLL) and Brier score \cite{brier1950verification}, two popular scores for computing how calibrated a binary classifier is. Formally, $NLL = \frac{1}{|Test|}\sum_{(\vertex{h}, \relation{r}, \vertex{t}),l\in Test} \function{softplus}(-l * \phi(\vertex{h}, \relation{r}, \vertex{t}))$ 
and 
$Brier = \frac{1}{|Test|}\sum_{(\vertex{h}, \relation{r}, \vertex{t}),l\in Test}
    (\sigma(\phi(\vertex{h}, \relation{r}, \vertex{t}))-l)^2$
where the equation for NLL assumes $l\in\{-1, +1\}$ and the equation for Brier score assumes $l\in\{0, 1\}$. For completeness sake, we also report AUC.

To measure the performance on WN18AM and FB15k-237AM, for each triple $(\vertex{h}, \relation{r}, \vertex{t})$ in the test set, we create two queries: $(?, \relation{r}, \vertex{t})$ and $(\vertex{h}, \relation{r}, ?)$. For each query, we use the model to rank all entities in $\vertices{E}$. Let $\kappa_{\vertex{h}, (?, \relation{r}, \vertex{t})}$ represent the ranking assigned to $\vertex{h}$ for query $(?, \relation{r}, \vertex{t})$ and $\kappa_{\vertex{t}, (\vertex{h}, \relation{r}, ?)}$ represent the ranking assigned to $\vertex{t}$ for query $(\vertex{h}, \relation{r}, ?)$. We compute \emph{mean reciprocal rank (MRR)} as the mean of the reciprocal of the ranks assigned to the correct entities and $hit@k$ as the fraction of queries for which the correct entity is ranked among the top $k$ entities. More formally, $MRR=\frac{1}{2*|Test|}\sum_{(\vertex{h}, \relation{r}, \vertex{t})\in Test} (\frac{1}{\kappa_{\vertex{h}, (?, \relation{r}, \vertex{t})}}+\frac{1}{\kappa_{\vertex{t}, (\vertex{h}, \relation{r}, ?)}})$
and
$Hit@k=\frac{1}{2*|Test|}\sum_{(\vertex{h}, \relation{r}, \vertex{t})\in Test} (\mathbbm{1}_{\kappa_{\vertex{h}, (?, \relation{r}, \vertex{t})}\leq k}+\mathbbm{1}_{\kappa_{\vertex{t}, (\vertex{h}, \relation{r}, ?)}\leq k})$
where $\mathbbm{1}_{condition}$ is $1$ if \emph{condition} holds and $0$ otherwise. The above setting is known as the \emph{raw} setting. \citet{bordes2013translating} advocated for a \emph{filtered} setting where MRR and Hit@k are computed similarly except that the ranking $\kappa_{\vertex{h}, (?, \relation{r}, \vertex{t})}$ is only computed among those entities $\vertex{e}$ where $(\vertex{e}, \relation{r}, \vertex{t})\not\in Train\cup Validation \cup Test$ (and similarly for $\kappa_{\vertex{t}, (\vertex{h}, \relation{r}, ?)}$). 

\textbf{Implementation:} 
To provide a fair comparison, we implemented our models and baselines in the same framework and measured their performance under fair experimental setup. The implementation is in PyTorch \cite{paszke2017automatic} and AdaGrad \cite{duchi2011adaptive} is used as the optimizer. We selected the best hyperparameters using grid search with values selected as follows: learning rate $\in\{10^{-1}, 10^{-2}, 10^{-3}, 10^{-4}\}$, $\lambda$ (in Equation~\ref{eq:loss-reg}) $\in\{10^{-1}, 10^{-3}\}$, L2 regularization $\in\{10^{-1}, 10^{-2}, 10^{-3}\}$, embedding size $\in\{100, 200\}$, dropout probability $\in\{0.0, 0.4\}$, negative ratio $\in\{1, 10\}$, and $\psi$ (in Equation~\ref{eq:reg}) $\in\{-3, -2, -1, 0\}$. We fix $I$ (introduced in the practical consideration paragraph of Section~\ref{sec:method}) to $5$ and $p$ (in Equation~\eqref{eq:reg}) to $1$. Note that the goal of our experiments is to show the merit of the proposed regularization compared to negative sampling as opposed to beating the state-of-the-art on a benchmark which requires much larger embedding dimensions, much larger grid for hyperparameter optimization, and adding several model-independent tricks \cite{kadlec2017knowledge,Ruffinelli2020you}.
For the experiments on WN11 and FB13, we selected the best hyperparameter setting and best epoch based on the NLL on the validation set. For the experiments on WN18AM and FB15k-237AM, we selected the best hyperparameter setting and best epoch based on the validation filtered MRR. The source code is available at \url{https://github.com/BorealisAI/StayPositive}.


\end{document}